\documentclass[11pt,a4paper]{article}
\usepackage[margin=25mm]{geometry}
\usepackage{lmodern}
\usepackage[sc,noBBpl]{mathpazo}
\usepackage[authoryear,round]{natbib}
\setcitestyle{citesep={;},aysep={,},yysep={;}}
\usepackage[T1]{fontenc}
\usepackage[utf8]{inputenc}
\usepackage{microtype}
\usepackage[table]{xcolor}
\usepackage{graphicx}
\usepackage{wrapfig}
\usepackage{needspace}
\usepackage{booktabs,makecell,array,tabularx}
\usepackage{amsmath,amssymb,mathtools}
\usepackage{pifont,enumitem}
\usepackage{caption}
\usepackage{listings}
\usepackage[most]{tcolorbox}
\usepackage{placeins}
\usepackage{hyperref}
\usepackage{xurl}
\definecolor{citeblue}{rgb}{0,0.08,0.45}
\definecolor{linkred}{HTML}{990000}
\hypersetup{
  colorlinks=true,allcolors=black,citecolor=citeblue,linkcolor=linkred,
  pdfauthor={Liangyu Teng, Yicheng Ding, Jing Liu, Hengsong Liu, Juncen Guo, Hongru Li, Jingyu Zhang, Liang Song},
  pdftitle={Marginal Response Surface Elicitation for Zero-Label Tabular Learning},
  pdfkeywords={Large language models, tabular learning, zero-label prediction}
}
\newcommand{\finishwrap}{\par
  \ifnum\value{WF@wrappedlines}>1
    \vspace{\dimexpr\baselineskip*(\value{WF@wrappedlines}-1)\relax}\fi
  \WFclear
}

\makeatletter
\renewcommand{\@seccntformat}[1]{\csname the#1\endcsname.\hspace{0.5em}}
\renewcommand{\section}{\@startsection{section}{1}{\z@}{-2.5ex plus -0.5ex minus -0.2ex}{1ex plus 0.2ex}{\normalfont\fontsize{13}{16}\selectfont\bfseries}}
\renewcommand{\subsection}{\@startsection{subsection}{2}{\z@}{-2ex plus -0.5ex minus -0.2ex}{0.8ex plus 0.2ex}{\normalfont\fontsize{12}{15}\selectfont\bfseries}}
\renewcommand{\subsubsection}{\@startsection{subsubsection}{3}{\z@}{-1.5ex plus -0.5ex minus -0.2ex}{0.6ex plus 0.2ex}{\normalfont\normalsize\bfseries}}
\renewcommand{\paragraph}{\@startsection{paragraph}{4}{\z@}{1.25ex plus 0.4ex minus 0.2ex}{-1em}{\normalfont\normalsize\bfseries}}
\makeatother

\definecolor{abstractcream}{HTML}{F7F0E3}
\renewenvironment{abstract}{\begin{tcolorbox}[enhanced,breakable,colback=abstractcream,colframe=abstractcream,
    boxrule=0pt,arc=5pt,boxsep=0pt,left=14pt,right=14pt,top=12pt,bottom=12pt,
    before skip=6pt,after skip=12pt]
  \normalfont\normalsize\setlength{\parindent}{0pt}\ignorespaces
}{\end{tcolorbox}}

\definecolor{marspromptteal}{RGB}{128,192,192}
\newtcolorbox{marspromptbox}[1]{enhanced,breakable,title={#1},colframe=marspromptteal,colback=white,
  colbacktitle=marspromptteal,coltitle=black,fonttitle=\bfseries,
  boxrule=0.8pt,arc=3pt,left=10pt,right=10pt,top=8pt,bottom=8pt,
  before skip=10pt,after skip=12pt
}
\lstdefinestyle{marsprompt}{basicstyle=\ttfamily\small,breaklines=true,breakatwhitespace=true,
  columns=fullflexible,keepspaces=true,showstringspaces=false,
  postbreak=\mbox{\textcolor{gray}{$\hookrightarrow$}\space},
  aboveskip=5pt,belowskip=8pt
}

\makeatletter
\renewcommand{\@maketitle}{\begingroup
  \setlength{\parindent}{0pt}\setlength{\parskip}{0pt}
  \vspace*{9pt}\nointerlineskip
  {\centering\fontsize{16}{19}\selectfont\bfseries\@title\par}
  \vspace{12pt}
  {\centering\normalsize\bfseries Fudan Institute on Networking Systems of AI\par}
  \vspace{18pt}
  \endgroup
}
\makeatother

\title{Marginal Response Surface Elicitation for Zero-Label Tabular Learning}
\newcommand{\paperauthors}{Liangyu~Teng\textsuperscript{1,2},
  Yicheng~Ding\textsuperscript{1},
  Jing~Liu\textsuperscript{3},
  Hengsong~Liu\textsuperscript{1,2},
  Juncen~Guo\textsuperscript{1,2},
  Hongru~Li\textsuperscript{1,2},
  Jingyu~Zhang\textsuperscript{1,2},
  Liang~Song\textsuperscript{1,2}}
\author{\paperauthors}
\date{}
\begin{document}
\maketitle
\begin{abstract}
Tabular learning uses structured data to predict target outcomes. Traditionally, this process has relied on labeled data. However, large language models (LLMs) can be used to elicit domain priors based on the task description and feature semantics, thereby enabling predictions without labeled data. We propose \textbf{Ma}rginal \textbf{R}esponse \textbf{S}urface Elicitation (MARS), a method that transforms feature-level LLM priors into a reusable, zero-shot tabular classifier. To construct this classifier, MARS selects representative values for each feature from unlabeled data and prompts the LLM to provide corresponding class support scores and feature weights. It then aggregates multiple responses using the median to construct feature response functions, and makes predictions through their weighted sum without further LLM queries. Across eight tabular benchmark tasks, MARS achieves the highest average AUC and AP, outperforming direct prompting by 1.97 and 6.21 percentage points respectively, while substantially reducing end-to-end costs. Evaluations with LLMs of different sizes further demonstrate its predictive advantage over direct prompting.

\end{abstract}

\section{Introduction}
\label{sec:intro}

Tabular learning utilizes structured information recorded in tables to predict target outcomes, finding broad applications across customer analytics, credit assessment, and business operations~\citep{borisov2024survey,hollmann2025tabpfn,gardner2024tabula}. Traditional prediction pipelines rely heavily on labeled samples, using supervised training to learn the mapping relationships between features and labels~\citep{grinsztajn2022trees,gorishniy2025tabm,erickson2025tabarena}. However, in many real-world business scenarios, the need for predictions often arises before actual business labels are available. For example, when a company launches a new product, it already has access to certain customer attributes and account information but needs to determine which customers to contact first before subscription results are available. This has given rise to the need for zero-label tabular prediction---that is, building a predictor using unlabeled records when the prediction target and feature meanings are known.

The knowledge accumulated by large language models (LLMs) during pretraining can be combined with task objectives to provide priors for label-scarce tabular prediction~\citep{bordt2024elephants,capstick2025autoelicit}. These priors are injected into the prediction pipeline in different ways: TabLLM~\citep{hegselmann2023tabllm} converts tabular records into text for direct prediction, CLLM~\citep{seedat2024cllm} and SERSAL~\citep{yan2025sersal} generate synthetic data and pseudo-labels to train the model, FeatLLM~\citep{han2024featllm} generates rules and converts them into features to train a simple model, and LLM-Trees~\citep{knauer2025llmtrees} and ProtoLLM~\citep{ma2025prototype} generate decision trees and prototypes to make predictions.

\Needspace{29\baselineskip}
\begin{wrapfigure}{r}{0.5\textwidth}
  \centering
  \captionsetup{justification=raggedright,singlelinecheck=false}
  \includegraphics[width=\linewidth]{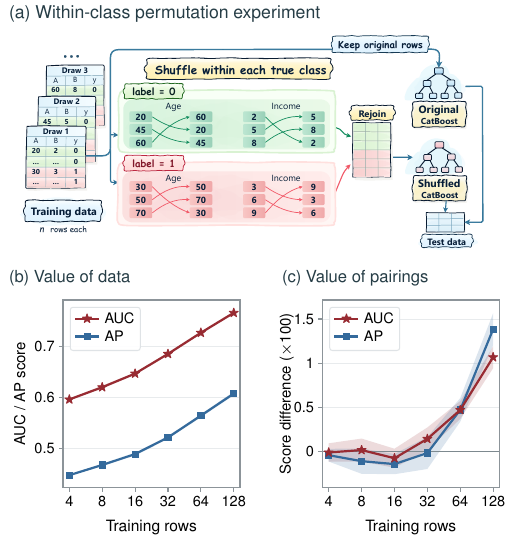}
\caption{Cross-feature couplings in few-shot learning. (a) Within-class column permutation preserves feature--class associations while disrupting original pairings. (b) CatBoost performance with increasing training samples. (c) Gains from retaining original pairings (original - shuffled, $\times100$). Shaded regions indicate $\pm1$ standard error across 16 seeds.}
  \label{fig:motivation}
\end{wrapfigure}

These methods raise a more fundamental question: What information can be leveraged in a few-shot setting? We distinguish between two types of information: \ding{182}~the relationship between the value of a single feature and the class, and \ding{183}~the additional information provided by cross-feature couplings. To guide which type of information to elicit from LLMs for zero-label prediction, we conduct a within-class column-wise permutation experiment~\citep{ojala2010permutation} on real-world datasets with labels. As illustrated in Figure~\ref{fig:motivation}, we independently permute the values of each feature within each class, thereby preserving the marginal distribution of each feature within each class, but disrupting the original cross-feature pairings. We then train identical CatBoost models separately on the original and permuted data, and compare their performance on the same test set. We find that the average performance difference is close to zero for 4--16 training samples, suggesting that the model gains little additional predictive benefit from cross-feature couplings in this regime. This suggests prioritizing the relationship between individual feature values and the class when eliciting LLM priors, as priors on cross-feature relationships may be noisy in the absence of labels.

Based on this intuition, we propose \textbf{Ma}rginal \textbf{R}esponse \textbf{S}urface Elicitation (MARS), a method that transforms the domain priors of LLMs into a reusable zero-label predictor. MARS selects representative values for each feature from unlabeled data and, combining task descriptions with feature semantics, obtains class support scores for each value and weights for each feature. Subsequently, it constructs feature response functions through median aggregation and employs their weighted sum for prediction. The entire process requires no ground-truth labels or local model training, and predicting new samples does not require calling the LLM again.

\finishwrap

\begingroup
\emergencystretch=1em
Comprehensive experiments across tabular benchmarks and LLMs of different sizes show that MARS is \ding{182}~\textbf{powerful}: MARS achieves the best average AUC and AP among all baselines, including direct LLM prompting; \ding{183}~\textbf{efficient}: MARS does not require any LLM API calls during inference, with end-to-end API call costs amounting to only 9.6\% of ProtoLLM; \ding{184}~\textbf{transferable}: MARS achieves better average AUC and AP than direct prompting on both Qwen3.5-4B and Qwen3.5-9B. In summary, our contributions are as follows:\par
\endgroup

\begin{itemize}[leftmargin=1em,labelindent=0pt,labelwidth=0.5em,labelsep=0.5em]
\item \textbf{Mechanistic Analysis:} The average benefit of within-class column permutation is close to zero in the few-shot regime, providing direct empirical support for the hypothesis that the LLM prior is informative at the feature level, but not at the instance level.
\item \textbf{Method Design:} We propose MARS, which converts feature-level LLM priors into response functions through median aggregation and combines them via weighted summation into a reusable predictor requiring no ground-truth labels, local model training, or test-time LLM queries.
\item \begin{samepage}\textbf{Experimental Evaluation:} On eight benchmark tasks, MARS achieves 1.97/6.21 higher average AUC/AP than direct prompting while largely reducing end-to-end costs compared to other high-performing baselines; cross-model experiments further demonstrate the transferability of our method.\par
  \end{samepage}
\end{itemize}

\section{Related Work}
\label{sec:related}

\paragraph{Few-Shot Tabular Learning.}
Few-shot tabular learning aims to achieve strong predictive performance with limited labeled data. To alleviate label scarcity, pretraining methods learn transferable feature representations from unlabeled data and adapt them to target tasks using a small number of labeled examples~\citep{yoon2020vime,bahri2022scarf,nam2023stunt}. Cross-task pretraining instead learns general predictive capabilities across large collections of tabular tasks and uses a few labeled examples as context to predict on new tasks~\citep{hollmann2025tabpfn,grinsztajn2026tabpfn3,qu2026tabiclv2}. These approaches reduce target-task annotation requirements through pretraining, whereas we investigate how to leverage LLM domain knowledge to construct predictors without target-task labels.

\paragraph{LLM-Based Tabular Learning.}
LLMs can leverage task descriptions and feature semantics to apply domain knowledge acquired during pretraining to tabular prediction. Direct-inference methods represent tabular records as text and use LLMs to generate class predictions~\citep{hegselmann2023tabllm,slack2023tablet,gardner2024tabula}. Knowledge-transfer methods instead convert LLM knowledge into training data or feature representations for learning downstream predictors~\citep{seedat2024cllm,yan2025sersal,han2024featllm,shi2025latte}. Another line of work uses LLMs to construct or refine predictors, encoding prior knowledge as decision rules or class prototypes for subsequent local prediction~\citep{knauer2025llmtrees,ye2025llm,ma2025prototype}. MARS transforms LLM domain knowledge into feature response functions and their weights, combining their weighted outputs into a reusable zero-label predictor. It requires neither downstream model training nor test-time LLM queries.

\section{Method}
\label{sec:method}

\begin{figure}[!t]
  \centering
  \includegraphics[width=\textwidth]{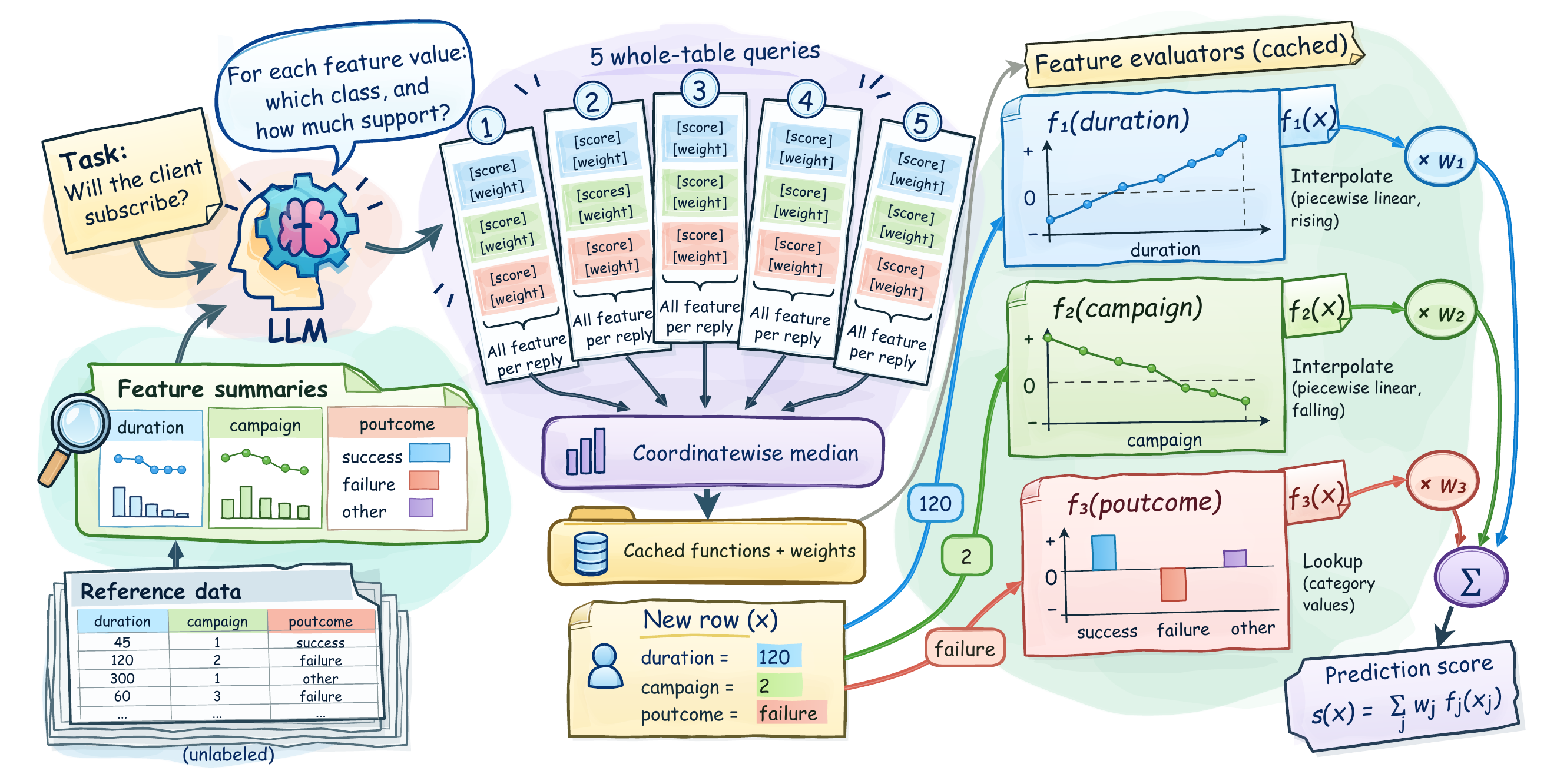}
\caption{Overview of MARS. MARS aggregates five LLM responses by the median and caches the resulting feature response functions and weights. New samples are scored by evaluating these functions at their feature values and summing the weighted responses.}
  \label{fig:pipeline}
\end{figure}

Figure~\ref{fig:pipeline} illustrates the overall pipeline of MARS. MARS first selects representative values for each feature from unlabeled reference data (Section~\ref{sec:feature-summaries}), and then prompts the LLM with the task description and feature semantics to obtain a support score for each value and a feature weight (Section~\ref{sec:response-elicitation}). Subsequently, MARS aggregates multiple responses to construct reusable feature response functions. When predicting new samples, it computes each feature's response via numerical interpolation or categorical lookup tables, and then aggregates the scores using the feature weights to obtain a final prediction score, without querying the LLM (Section~\ref{sec:response-prediction}).

\subsection{Representative Value Selection}
\label{sec:feature-summaries}

Let $U$ be the unlabeled reference table with $p$ features. For each feature $j$, MARS selects $m_j$ representative values $a_{j1},\ldots,a_{jm_j}$ as anchors, queries the LLM for their scores, and uses the scores to build the feature's response function.

Tabular features are either numerical or categorical. For a numerical feature, MARS computes the empirical quantiles at $\{0.05,0.20,0.35,0.50,0.65,0.80,0.95\}$ using the non-missing values in $U$ and removes duplicates to obtain at most seven anchors. For a categorical feature, MARS stores up to 20 of the most frequent values and their frequencies. The feature name, type, and anchors are then sent to the LLM together with the task description.

\subsection{Marginal Response Elicitation}
\label{sec:response-elicitation}

MARS combines the task and feature descriptions into a single prompt and formulates the binary classification task as a yes/no question where ``yes'' represents the positive class. We query the LLM $R=5$ times using the same prompt, covering all features each time, and ask the LLM to first explain the meaning of the positive class and provide typical examples in 1--2 sentences before scoring each feature.

In the $r$-th response, the LLM provides a support score $z_{j\ell}^{(r)}\in[-3,3]$ for each anchor $a_{j\ell}$ of feature $j$. Positive, negative, and zero values represent support for the positive class, negative class, and neutrality, respectively, while the absolute value indicates support strength. The prompt requires that all features use a uniform log-odds scale, where $+1$ and $-1$ approximately correspond to multiplying and dividing the odds of the positive class by $e$, respectively.

The LLM also assigns an importance score $b_j^{(r)}\in[0,10]$ to each feature, which measures its importance in distinguishing classes relative to other features; a score of 0 indicates that the feature is not important. Unlike the support scores assigned to individual values, this score is uniform across the entire feature and is subsequently used to calculate the global weight.

We request the LLM to return the importance scores of each feature and the support scores arranged in the order of the anchors in JSON format. The complete prompt is provided in Appendix~\ref{app:prompt}.

\subsection{Response Aggregation and Prediction}
\label{sec:response-prediction}

Let $\mathcal V_j$ be the set of response indices for feature $j$. MARS then aggregates the anchor support scores and feature importance scores using the median, to obtain the final anchor support score $\bar z_{j\ell}$ and the final feature weight $w_j$:
\begin{equation}
\begin{aligned}
\bar z_{j\ell}
&=\operatorname*{median}_{r\in\mathcal V_j}
\operatorname{clip}_{[-3,3]}\bigl(z_{j\ell}^{(r)}\bigr),\\
w_j
&=\frac{1}{10}\max\left(
0,\operatorname*{median}_{r\in\mathcal V_j}b_j^{(r)}
\right).
\end{aligned}
\label{eq:aggregate}
\end{equation}
The median is robust to outliers. If there are no responses for a feature, it is not used in the prediction.

MARS then constructs the feature response function $f_j$ using the anchors and the aggregated support scores. For numerical features, $f_j$ is a piecewise linear function defined as
\begin{equation}
f_j(v)=\bar z_{j\ell}
+\frac{v-a_{j\ell}}{a_{j,\ell+1}-a_{j\ell}}
\left(\bar z_{j,\ell+1}-\bar z_{j\ell}\right),
\label{eq:interpolation}
\end{equation}
for $a_{j\ell}\le v\le a_{j,\ell+1}$. For values outside the range of the anchors, the function is extended using the response at the closest anchor. For categorical features, $f_j$ is a lookup table defined as
\begin{equation}
f_j(v)=
\begin{cases}
\bar z_{j\ell}, & v=a_{j\ell},\quad \ell=1,\ldots,m_j,\\
0, & \text{otherwise}.
\end{cases}
\label{eq:categorical-response}
\end{equation}

Given a new instance $x=(x_1,\ldots,x_p)$, the MARS score is computed as
\begin{equation}
s(x)=\sum_{j:\,\mathcal V_j\neq\emptyset}w_j f_j(x_j).
\label{eq:score}
\end{equation}
A higher score indicates stronger support for the positive class. The resulting model is a nonlinear additive model, where each feature is modeled using a nonlinear function, and the nonlinear functions are combined additively. Note that the model is constructed using a finite number of queries to the LLM, and can be used to score new instances without further querying the LLM.

\section{Experiments}
\label{sec:results}

\begin{table}[!t]
\centering
\caption{AUC/AP ($\uparrow$) on eight tabular benchmarks. Best scores are \textbf{bold}; $\dagger$ denotes test-time LLM queries.}
\label{tab:main}
\fontsize{8}{10.2}\selectfont
\setlength{\tabcolsep}{1.5pt}
\renewcommand{\arraystretch}{1.0}
\begin{tabular*}{\textwidth}{@{\extracolsep{\fill}}lc*{9}{c}@{}}
\toprule
\textbf{Method} & \textbf{Labels} & \textbf{Bank} & \textbf{Blood} & \textbf{Credit-G} & \textbf{Diabetes} & \textbf{Heart} & \textbf{Cultivars} & \textbf{Myocardial} & \textbf{NHANES} & \textbf{Avg.} \\
\midrule
LogReg & 4 & .526/.162 & .282/.168 & .601/.796 & .545/.432 & .508/.553 & .584/.610 & .486/.230 & .544/.165 & .509/.389 \\
XGBoost & 4 & .500/.117 & .500/.240 & .500/.700 & .500/.351 & .500/.554 & .500/.500 & .500/.225 & .500/.160 & .500/.356 \\
CatBoost & 4 & .562/.166 & .364/.198 & .518/.729 & .411/.355 & .565/.636 & .638/.619 & .469/.257 & .569/.208 & .512/.396 \\
STUNT & 4 & .509/.123 & .427/.229 & .461/.690 & .731/.591 & .908/.921 & .480/.589 & .579/.309 & .425/.132 & .565/.448 \\
TabPFN-3 & 4 & .564/.169 & .402/.219 & .611/.793 & .484/.360 & .703/.693 & .633/.597 & .487/.285 & .531/.177 & .552/.412 \\
TabICLv2 & 4 & .633/.197 & .382/.217 & .546/.744 & .560/.411 & .750/.763 & .642/.623 & .523/.280 & .530/.190 & .571/.428 \\
TabFM & 4 & .396/.092 & .437/.239 & .576/.750 & .579/.371 & .506/.556 & .576/.568 & .417/.189 & .617/.233 & .513/.375 \\
DeLTa & 4 & .703/.252 & .357/.203 & .395/.624 & .552/.414 & .509/.574 & .534/.570 & .477/.218 & .452/.155 & .498/.376 \\
FeatLLM & 4 & .667/.226 & .411/.255 & .472/.693 & .760/.583 & .839/.861 & \textbf{.662}/.621 & .608/.334 & .585/.242 & .626/.477 \\
\midrule
Direct$^{\dagger}$ & 0 & .841/.463 & .683/.353 & .596/.773 & .808/.693 & .894/.885 & .593/.586 & .666/.320 & .690/\textbf{.303} & .721/.547 \\
TabuLa-8B$^{\dagger}$ & 0 & .808/.417 & .590/.310 & .419/.673 & .717/.612 & .767/.760 & .443/.466 & .623/.372 & .444/.148 & .601/.470 \\
LLM-Trees & 0 & .580/.213 & .612/.302 & .593/.744 & .733/.560 & .802/.766 & .537/.524 & .593/.283 & .598/.213 & .631/.451 \\
ProtoLLM & 0 & .820/.392 & \textbf{.725}/\textbf{.448} & \textbf{.660}/\textbf{.818} & \textbf{.870}/.757 & .881/.904 & .598/\textbf{.684} & .594/.318 & .687/.261 & .729/.573 \\
\midrule
\textbf{MARS (ours)} & 0 & \textbf{.884}/\textbf{.630} & .693/.427 & .610/.790 & .858/\textbf{.757} & \textbf{.927}/\textbf{.940} & .586/.604 & \textbf{.673}/\textbf{.443} & \textbf{.698}/.283 & \textbf{.741}/\textbf{.609} \\
\bottomrule
\end{tabular*}
\end{table}

\subsection{Experimental Setup}
\label{sec:setup}

\textbf{Tasks and Benchmarks.} Following FeatLLM~\citep{han2024featllm} and ProtoLLM~\citep{ma2025prototype}, we evaluate MARS on eight tabular benchmark tasks covering \textbf{financial services} (Bank~\citep{moro2011bank}, Credit-G~\citep{hofmann1994german}), \textbf{healthcare} (Blood~\citep{yeh2009rfm}, Diabetes~\citep{smith1988adap}, Heart~\citep{fedesoriano2021heart}, Myocardial~\citep{golovenkin2020trajectories}, NHANES~\citep{uci2019nhanes}), and \textbf{agriculture} (Cultivars~\citep{oliveira2023cultivars}). The evaluation metrics are ROC-AUC (AUC) and average precision (AP).

\textbf{Baselines.} We compare with three groups of baselines: (1) \textbf{supervised learning}, including LogReg, XGBoost~\citep{chen2016xgboost}, and CatBoost; (2) \textbf{few-shot tabular learning}, including STUNT~\citep{nam2023stunt}, TabPFN-3~\citep{grinsztajn2026tabpfn3}, TabICLv2~\citep{qu2026tabiclv2}, TabFM~\citep{tabfm2026software}, DeLTa~\citep{ye2025llm}, and FeatLLM~\citep{han2024featllm}; (3) \textbf{zero-label prediction}, including Direct (direct LLM prompting), TabuLa-8B~\citep{gardner2024tabula}, LLM-Trees~\citep{knauer2025llmtrees}, and ProtoLLM~\citep{ma2025prototype}. The first two groups of baselines are provided with 4 labeled examples (2 per class) for both training and model selection. The third group of baselines, as well as MARS, do not use any task-specific labels for prediction.

\textbf{Implementation Details.} In the main experiments, we use DeepSeek-V4-Flash-0731 with high-intensity thinking as the backbone LLM. We also use Qwen3.5-4B and Qwen3.5-9B to compare MARS with Direct, with native thinking, temperature=1.0, top-p=0.95, top-k=20, presence penalty=1.5. For a fair comparison, MARS uses the same unlabeled reference data as ProtoLLM, with $R=5$ queries executed per task. Appendices~\ref{app:data} and~\ref{app:settings} provide the fixed data protocol and implementation settings.

\Needspace{23\baselineskip}
\subsection{Predictive Performance}
\label{sec:matrix}
\label{sec:qwen}

\begin{wraptable}[15]{r}{0.5\textwidth}
\vspace{-1.5\baselineskip}
\centering
\captionsetup{justification=raggedright,singlelinecheck=false}
\caption{Eight-task ablation and backbone averages. Changes (pp) are relative to Direct with the same backbone. Best scores per backbone are \textbf{bold}.}
\label{tab:analysis}
\fontsize{8.5}{10.5}\selectfont
\definecolor{MARSTableGroup}{HTML}{E2EFF0}
\definecolor{MARSTableGain}{HTML}{A74F3D}
\definecolor{MARSTableLoss}{HTML}{346A9C}
\newcommand{\marsmetric}[2]{\makebox[23pt][r]{#1}\hspace{2pt}\makebox[29pt][l]{#2}}
\setlength{\tabcolsep}{2pt}
\renewcommand{\arraystretch}{1.0}
\begin{tabular*}{\linewidth}{@{\extracolsep{\fill}}lcc@{}}
\toprule
\textbf{Method} & \textbf{Avg. AUC} $\uparrow$ & \textbf{Avg. AP} $\uparrow$ \\
\specialrule{.4pt}{2pt}{0pt}
\rowcolor{MARSTableGroup}[0pt][0pt]
\multicolumn{3}{@{}c@{}}{\rule[-4pt]{0pt}{14pt}\raisebox{-2pt}{\includegraphics[height=10pt]{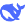}}\hspace{4pt}\textbf{DeepSeek-V4-Flash-0731}} \\
\specialrule{.4pt}{0pt}{0pt}
Direct & \marsmetric{.7213}{} & \marsmetric{.5471}{} \\
\textbf{MARS} & \marsmetric{\textbf{.7411}}{\textcolor{MARSTableGain}{$\uparrow 1.97$}} & \marsmetric{\textbf{.6091}}{\textcolor{MARSTableGain}{$\uparrow 6.21$}} \\
\hspace{.6em}w/o feature weights & \marsmetric{.7363}{\textcolor{MARSTableGain}{$\uparrow 1.50$}} & \marsmetric{.5966}{\textcolor{MARSTableGain}{$\uparrow 4.95$}} \\
\hspace{.6em}w/o response aggregation & \marsmetric{.7358}{\textcolor{MARSTableGain}{$\uparrow 1.45$}} & \marsmetric{.6002}{\textcolor{MARSTableGain}{$\uparrow 5.32$}} \\
\hspace{.6em}w/o support strength & \marsmetric{.7094}{\textcolor{MARSTableLoss}{$\downarrow 1.19$}} & \marsmetric{.5554}{\textcolor{MARSTableGain}{$\uparrow 0.83$}} \\
\specialrule{.4pt}{2pt}{0pt}
\rowcolor{MARSTableGroup}[0pt][0pt]
\multicolumn{3}{@{}c@{}}{\rule[-4pt]{0pt}{14pt}\raisebox{-2pt}{\includegraphics[height=10pt]{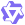}}\hspace{4pt}\textbf{Qwen3.5-4B}} \\
\specialrule{.4pt}{0pt}{0pt}
Direct & \marsmetric{.5745}{} & \marsmetric{.4366}{} \\
\textbf{MARS} & \marsmetric{\textbf{.6713}}{\textcolor{MARSTableGain}{$\uparrow 9.69$}} & \marsmetric{\textbf{.5257}}{\textcolor{MARSTableGain}{$\uparrow 8.91$}} \\
\specialrule{.4pt}{2pt}{0pt}
\rowcolor{MARSTableGroup}[0pt][0pt]
\multicolumn{3}{@{}c@{}}{\rule[-4pt]{0pt}{14pt}\raisebox{-2pt}{\includegraphics[height=10pt]{figures/backbone_icons/qwen.pdf}}\hspace{4pt}\textbf{Qwen3.5-9B}} \\
\specialrule{.4pt}{0pt}{0pt}
Direct & \marsmetric{.6364}{} & \marsmetric{.4794}{} \\
\textbf{MARS} & \marsmetric{\textbf{.6695}}{\textcolor{MARSTableGain}{$\uparrow 3.31$}} & \marsmetric{\textbf{.5148}}{\textcolor{MARSTableGain}{$\uparrow 3.54$}} \\
\bottomrule
\end{tabular*}
\end{wraptable}

As shown in Table~\ref{tab:main}, MARS achieves the best average AUC and AP across all eight tasks, outperforming Direct by 1.97 and 6.21 percentage points, and the strongest baseline ProtoLLM by 1.17 and 3.64 percentage points, respectively. Across both metrics, MARS surpasses Direct on seven tasks each, demonstrating that these gains are not confined to isolated tasks. Table~\ref{tab:analysis} further presents the results on backbones of different scales: on Qwen3.5-4B and Qwen3.5-9B, the average AUC/AP of MARS outperforms Direct with the corresponding backbones by 9.69/8.91 and 3.31/3.54 percentage points, respectively, indicating that feature-level prior elicitation sustains its predictive advantage across both model scales.

\finishwrap

\Needspace{17\baselineskip}
\subsection{Efficiency and Sensitivity Analysis}
\label{sec:cost}
\label{sec:query}

\begin{wrapfigure}[12]{r}{0.5\textwidth}
\vspace{-1.5\baselineskip}
\centering
\captionsetup{justification=raggedright,singlelinecheck=false}
\includegraphics[width=\linewidth]{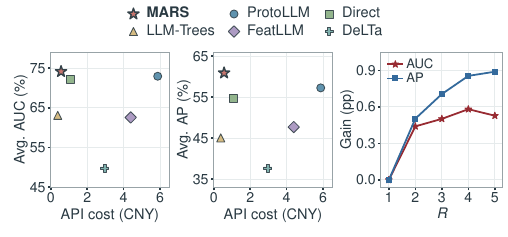}\par
\caption{Cost--performance trade-offs and response sensitivity. (a,b) Mean cost and performance; upper left is better. (c) AUC/AP gains over one LLM response ($R=1$).}
\label{fig:cost}
\end{wrapfigure}

Figure~\ref{fig:cost}(a,b) compares the average predictive performance and API costs of different methods across the eight tasks. We account for actual token consumption based on the official DeepSeek API peak-hour pricing. The average task cost of MARS is 0.564 yuan, amounting to only 9.6\% of ProtoLLM and lower than Direct's 1.065 yuan, while achieving higher average AUC and AP. The LLM queries of MARS are concentrated exclusively in the predictor construction stage; once constructed, new samples can be processed via local computation without additional LLM calls.

Figure~\ref{fig:cost}(c) further analyzes the impact of the number of responses $R$. For each $R$, we evaluate all subsets of size $R$ from the existing five responses and report the average performance. Increasing the number of responses from 1 to 5 increases the mean AUC and AP by 0.53 and 0.89 percentage points, respectively, with diminishing gains for larger $R$.

\finishwrap

\subsection{Ablation Study}
\label{sec:analysis}
\label{sec:ablation}

Table~\ref{tab:analysis} examines the roles of feature weights, multi-response aggregation, and support strength by uniformly setting feature weights $w_j$ to $1$, using only a single response ($R=1$), and replacing aggregated anchor scores with their signs ($-1/0/+1$), respectively. In the support-strength ablation, sign replacement is applied before interpolation or lookup, with feature weights unchanged. Compared to the full MARS, all three settings degrade the average AUC and AP. Removing support strength causes the most pronounced drops, demonstrating that the degree of support conveys predictive information beyond the direction of support.

\Needspace{17\baselineskip}
\subsection{Case Study}
\label{sec:case}

\begin{wrapfigure}{r}{0.5\textwidth}
\centering
\captionsetup{justification=raggedright,singlelinecheck=false}
\includegraphics[width=\linewidth]{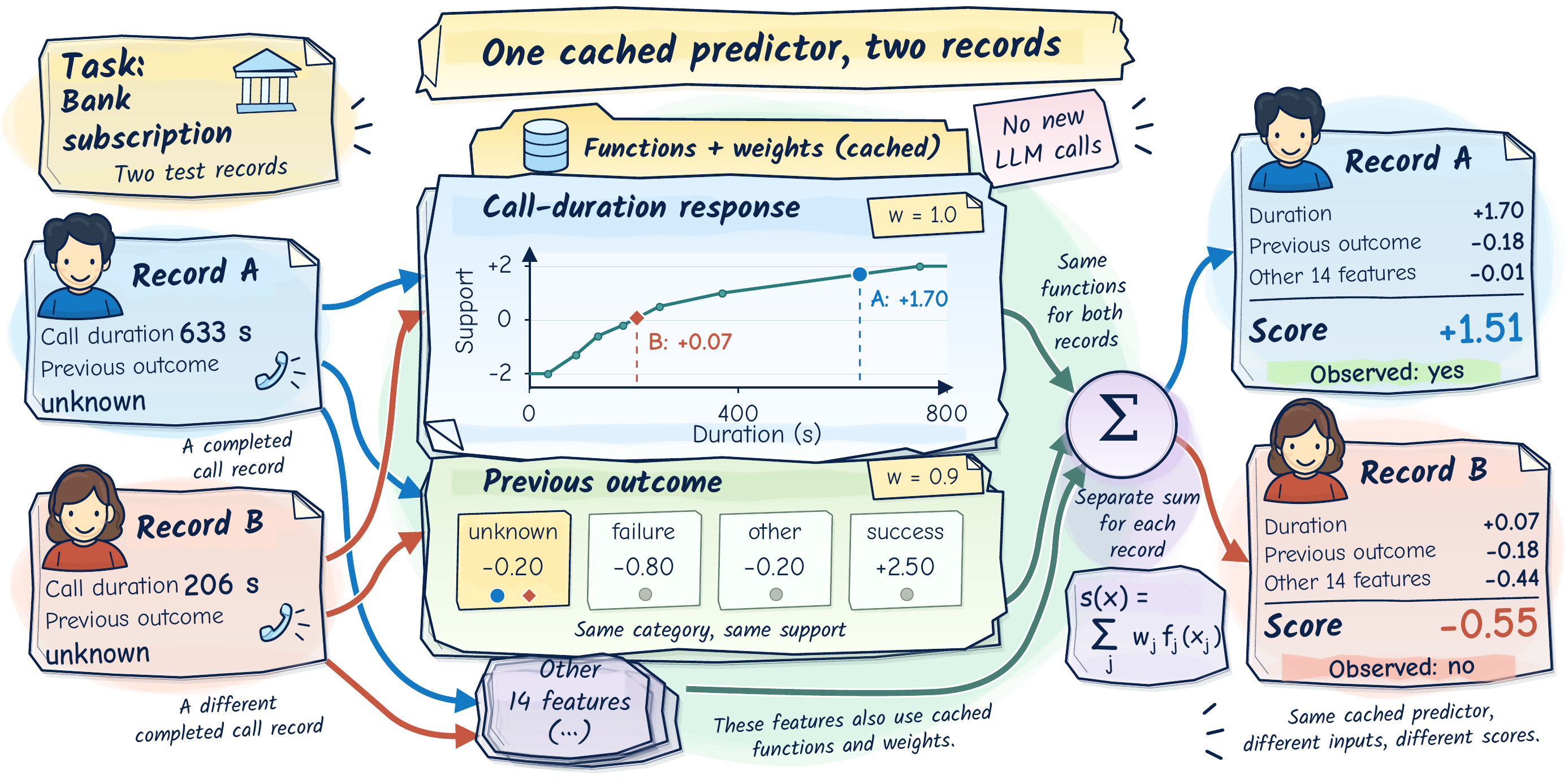}\par
\caption{Case study on Bank. Two records are scored using the same cached response functions and feature weights, without further LLM queries.}
\label{fig:case}
\end{wrapfigure}

Figure~\ref{fig:case} illustrates how MARS translates LLM domain priors into feature-level predictive contributions. The same response function assigns support scores of 1.70 and 0.07 to call durations of 633 and 206 seconds, respectively, capturing how specific values affect positive-class support strength. The shared previous outcome, \textit{unknown}, contributes the same weighted value of $-0.18$ to both records. These positive and negative contributions combine with those of the other features to produce different final scores. By modeling the direction and strength of value-specific support and combining responses with feature weights, MARS turns the LLM's semantic judgments into decomposable predictive contributions.

\finishwrap

\section{Conclusion}
\label{sec:conclusion}

We propose MARS for zero-label tabular prediction, transforming feature-level LLM priors into response functions and combining them through weighted summation to construct a reusable predictor. Experiments on eight benchmark tasks show that MARS achieves the highest average AUC and AP among the compared methods while remaining cost-efficient. Future work will explore how to reliably incorporate cross-feature interactions to further improve predictive performance.

\FloatBarrier

\Needspace{10\baselineskip}
\section*{Authors and Affiliations}
\begingroup
\small\setlength{\parindent}{0pt}\setlength{\parskip}{0pt}
\raggedright
\renewcommand{\thefootnote}{\fnsymbol{footnote}}
\paperauthors\footnote[1]{Corresponding author}\par
\vspace{9pt}
\renewcommand{\arraystretch}{1.12}
\begin{tabularx}{\linewidth}{@{}r@{\hspace{0.65em}}>{\raggedright\arraybackslash}X@{}}
\textsuperscript{1} & College of Intelligent Robotics and Advanced Manufacturing, Fudan University, Shanghai, China\\[4pt]
\textsuperscript{2} & International Networking Systems of Artificial Intelligence Limited, Macau\\[4pt]
\textsuperscript{3} & College of Future Information Technology, Fudan University, Shanghai, China
\end{tabularx}\par
\endgroup

\urlstyle{same}
\bibliographystyle{plainnat}
\bibliography{references}

\clearpage
\appendix
\section{Prompting Protocol}
\label{app:prompt}

MARS uses the following system and user messages for all three backbones. The user message substitutes the binary task question for \texttt{\{question\}} and the feature summaries for \texttt{\{fields\}}.

\begin{marspromptbox}{MARS Prompt Template}
\textbf{System message}
\begin{lstlisting}[style=marsprompt]
You are a domain expert. Using only your knowledge of the domain, you quantify how each field of a table relates to the target described in the question. You have no labelled examples. Answer with a single JSON object and nothing else.
\end{lstlisting}
\textbf{User message}
\begin{lstlisting}[style=marsprompt]
Task question: {question}
Answering "yes" is the positive class.

Fields, each with reference points computed from unlabelled rows (seven quantiles for numeric fields; levels with relative frequencies for categorical fields): {fields}

First, in one or two sentences, restate what the positive class means and describe a typical positive case.

Then, for each field, give the log-odds of the positive class at each of its reference points, on a common scale: 0.0 is no evidence, +1.0 roughly multiplies the odds by e, -1.0 roughly divides them by e; keep values within [-3, 3]. Values need not follow a straight line across the reference points. Keep the sign of each value consistent with your description: a value typical of a positive case gets a positive log-odds.

For each field also give discriminative_strength in [0, 10]: how strongly it separates the classes relative to the other fields; 0 means irrelevant.

Answer with exactly this JSON object:
{"positive_class_means": "<one sentence>",
 "typical_positive_case": "<one sentence>",
 "fields": [
  {"name": "<field name exactly as given>",
   "discriminative_strength": <0-10>,
   "logit_at_points": [<one number per reference point, same order>]}
 ]}
\end{lstlisting}
\end{marspromptbox}

For a numerical feature, its summary gives the feature name, type, and ordered reference values. For a categorical feature, it gives the retained levels and their relative frequencies. Numerical anchors are rounded to six decimal places before duplicate removal; their textual values use general numeric formatting. Categorical frequencies are displayed as integer percentages. The prompt's \texttt{logit\_at\_points} and \texttt{discriminative\_strength} correspond to the anchor support scores and feature importance in Section~\ref{sec:response-elicitation}.

\clearpage
\section{Benchmark and Evaluation Protocol}
\label{app:data}

\subsection{Datasets and Prediction Tasks}

We evaluate MARS on eight binary classification tasks spanning financial services, healthcare, and agriculture. Table~\ref{tab:dataset-details} summarizes the dataset sizes and feature types. Bank, Blood, Credit-G, Diabetes, Heart, and Myocardial use the processed datasets released with FeatLLM~\citep{han2024featllm}, retaining their feature names, types, and task questions and mapping \texttt{no}/\texttt{yes} to 0/1. Cultivars and NHANES use their public dataset releases. The prediction tasks and dataset-specific preparation are described below.

\begin{table}[!ht]
\centering
\caption{Benchmark sizes. Num./Cat. counts numerical and categorical input features; the target column is excluded.}
\label{tab:dataset-details}
\small
\setlength{\tabcolsep}{10pt}
\begin{tabular*}{\linewidth}{@{\extracolsep{\fill}}lrrrr@{}}
\toprule
Task & Rows & Features & Num./Cat. & Test rows \\
\midrule
Bank & 45,211 & 16 & 7/9 & 800 \\
Blood & 748 & 4 & 4/0 & 150 \\
Credit-G & 1,000 & 20 & 7/13 & 200 \\
Diabetes & 768 & 8 & 8/0 & 154 \\
Heart & 918 & 11 & 6/5 & 184 \\
Cultivars & 320 & 10 & 7/3 & 64 \\
Myocardial & 686 & 91 & 7/84 & 138 \\
NHANES & 2,278 & 7 & 4/3 & 456 \\
\bottomrule
\end{tabular*}
\end{table}

\begin{enumerate}[leftmargin=*,itemsep=3pt,topsep=6pt,parsep=0pt]
\item \textbf{Bank.} A telephone-marketing dataset~\citep{moro2011bank} for predicting whether a customer subscribes to a term deposit using customer background and contact records.

\item \textbf{Blood.} A blood-donation dataset~\citep{yeh2009rfm} for predicting whether an individual donates blood based on past donation records, including donation recency, frequency, and cumulative volume.

\item \textbf{Credit-G.} A credit-assessment dataset~\citep{hofmann1994german} for classifying individual credit risk using account status, credit history, and loan information.

\item \textbf{Diabetes.} A diabetes classification dataset~\citep{smith1988adap} that uses clinical features such as glucose level, body mass index, and age to predict whether a patient has diabetes.

\item \textbf{Heart.} A heart-disease classification dataset~\citep{fedesoriano2021heart} that combines clinical features such as chest pain type, blood pressure, and heart rate to predict the presence of heart disease.

\item \textbf{Cultivars.} A dataset of soybean cultivars and field observations~\citep{oliveira2023cultivars} for predicting whether grain yield (\texttt{GY}) exceeds the fixed dataset median of 3397.276724 kg/ha using cultivar information and plant traits. Grain yield itself is excluded from the inputs; \texttt{Season}, \texttt{Cultivar}, and \texttt{Repetition} are categorical features. We split individual observations rather than holding out entire cultivars, so the same cultivar may appear in both the training and test partitions.

\item \textbf{Myocardial.} Clinical records of myocardial infarction patients~\citep{golovenkin2020trajectories}. We use the chronic-heart-failure classification task, with patient history and clinical measurements as inputs, retaining the released history feature \texttt{ZSN\_A}.

\item \textbf{NHANES.} A health and nutrition survey dataset~\citep{uci2019nhanes} for predicting whether a participant belongs to the age group of 65 years or older from questionnaire responses, physical examination results, and biochemical measurements. Age (\texttt{RIDAGEYR}), the age-group label (\texttt{age\_group}), and the participant identifier (\texttt{SEQN}) are excluded from the inputs.
\end{enumerate}

\subsection{Splits, Label Access, and Metrics}

We use an 80/20 stratified split with random seed 0. From the training partition, two examples per class are reserved for the four-label comparators. The remaining training features form the unlabeled reference pool. The same support examples and test records are shared by the methods. For test partitions larger than 800 rows, we sample approximately 800 rows proportionally by class with seed 0; unused test records are not moved into the reference pool. Table~\ref{tab:dataset-details} gives the resulting evaluation sizes, totaling 2,146 records.

Labels are used by the evaluator to define the split, select the four comparator examples, and compute metrics. MARS receives only the task question and reference features. Its prompt, anchor selection, aggregation, and prediction do not access support or test labels. The comparison uses no additional labeled validation set.

We compute ROC-AUC and average precision from the continuous prediction scores and average task-level metrics with equal task weights. The main table reports this fixed split. The five elicitation responses are repeated LLM requests, not five data-split seeds. For response-count sensitivity, every subset of size $R$ from the five cached responses is evaluated, first averaging over subsets within each task and then over tasks. The single-response ablation averages the five one-response models.

\FloatBarrier
\section{Hyperparameters and Implementation Details}
\label{app:settings}

\subsection{MARS Construction Settings}

The construction settings in Table~\ref{tab:mars-settings} are shared by the eight tasks and three backbones. MARS constructs response functions by aggregation and interpolation and elicits feature weights from the LLM, without a task-specific optimizer, learning rate, or training schedule. For each feature, we aggregate valid replies whose score vectors contain exactly one finite value per anchor.

\begin{table}[!ht]
\centering
\caption{MARS construction settings used in the reported experiments.}
\label{tab:mars-settings}
\small
\renewcommand{\arraystretch}{1.12}
\begin{tabularx}{\linewidth}{@{}p{0.38\linewidth}X@{}}
\toprule
Setting & Value \\
\midrule
Task-specific labeled examples & 0 \\
LLM responses per task & $R=5$, all retained features per response \\
Numerical quantiles & $0.05, 0.20, 0.35, 0.50, 0.65, 0.80, 0.95$ \\
Numerical anchor preparation & Round to six decimals, remove duplicates; omit features with fewer than two anchors \\
Categorical levels & Up to 20 most frequent levels \\
Anchor score clipping & $[-3,3]$, before aggregation \\
Feature importance requested & $[0,10]$ \\
Aggregation & Coordinatewise median over valid feature responses \\
Feature weight & $\max(0,\operatorname{median} b_j^{(r)})/10$ \\
Numerical prediction & Piecewise linear interpolation, endpoint extension \\
Categorical prediction & Lookup; unseen levels contribute 0 \\
\bottomrule
\end{tabularx}
\end{table}

\subsection{LLM Generation Settings}

Table~\ref{tab:generation-settings} records the generation settings for MARS. DeepSeek uses the API model identifier \texttt{deepseek-v4-flash}, recorded as DeepSeek-V4-Flash-0731 in the experiment. Thinking is enabled with high reasoning effort and JSON-object output. Temperature and top-$p$ were not explicitly set in these requests. Qwen3.5-4B and Qwen3.5-9B use their native thinking template with the same sampling settings as each other.

\begin{table}[!ht]
\centering
\caption{MARS generation settings. ``Not set'' means the request contains no override for that parameter.}
\label{tab:generation-settings}
\small
\renewcommand{\arraystretch}{1.12}
\begin{tabularx}{\linewidth}{@{}p{0.34\linewidth}XX@{}}
\toprule
Parameter & DeepSeek-V4-Flash & Qwen3.5-4B / 9B \\
\midrule
Thinking & Enabled; high effort & Native template enabled \\
Maximum output tokens & 32,768 & 32,768 \\
Temperature & Not set & 1.0 \\
Top-$p$ & Not set & 0.95 \\
Top-$k$ & Not set & 20 \\
Min-$p$ & Not set & 0.0 \\
Presence penalty & Not set & 1.5 \\
Repetition penalty & Not set & 1.0 \\
Frequency penalty & Not set & 0.0 \\
\bottomrule
\end{tabularx}
\end{table}

\Needspace{7\baselineskip}
Qwen inference uses bfloat16, tensor parallel size 1, engine seed 0, a maximum context length of 81,920 tokens, and GPU-memory utilization 0.92. The engine allows at most 8 sequences and 4,096 batched tokens, with chunked prefill and prefix caching enabled.

\FloatBarrier

\end{document}